\documentclass[11pt,a4paper]{article}
\usepackage[hyperref]{acl2020}
\usepackage{times}
\usepackage{latexsym}
\usepackage{xcolor}
\usepackage{pgfplots}
\usepackage{graphics}
\usepackage{clrscode3e}
\usepackage{booktabs}
\usepackage{multirow}
\usepackage{amsmath}
\usepackage{amsthm}
\usepackage{amssymb}
\usepackage{makecell}
\usepackage{tikz}
\usepackage{float}
\usetikzlibrary{arrows,shapes,positioning}
\usepackage{tikz-qtree}

\usepackage{microtype}

\aclfinalcopy 

\newcommand{\word}[1]{\emph{#1}}
\newcommand{\tech}[1]{\emph{#1}}
\renewcommand{\textit}[1]{\emph{#1}}

\newcommand{\secref}[1]{Section~\ref{#1}}
\newcommand{\appref}[1]{Appendix~\ref{#1}}

\newcommand{\Figref}[1]{Figure~\ref{#1}}
\newcommand{\figref}[1]{Figure~\ref{#1}}

\newcommand{\tabref}[1]{Table~\ref{#1}}

\theoremstyle{definition}

\newcommand{\MoNLI}{\textrm{MoNLI}}
\newcommand{\NMoNLI}{\textrm{NMoNLI}}
\newcommand{\PMoNLI}{\textrm{PMoNLI}}
\newcommand{\BERT}{\text{BERT}}

\newcommand{\CLS}{\texttt{[CLS]}}
\newcommand{\SEP}{\texttt{[SEP]}}

\newcommand{\varfunc}{\id{lexrel}}
\newcommand{\outfunc}{\proc{Infer}}
\newcommand{\interfunc}[3]{\outfunc_{#3(#1) \to #3(#2)}(#1)}
\newcommand{\interv}[3]{\BERT_{#3(#1) \to #3(#2)}(#1)}
\newcommand{\lexvar}{\id{lexrel}}
\newcommand{\entailrelations}{\{\sqsupset, \sqsubset\}}
\newcommand{\bertmap}{L}

\newcommand{\lexrelproc}{\proc{get-lex-rel}}

\title{Neural Natural Language Inference Models Partially Embed Theories of Lexical Entailment and Negation}

\author{Atticus Geiger \\
  Stanford University \\
  \texttt{atticusg@stanford.edu} \\\And
  Kyle Richardson \\
  Allen Institute for AI\\
  \texttt{kyler@allenai.org} \\\And
  Christopher Potts \\
  Stanford University \\
  \texttt{cgpotts@stanford.edu}
}

\date{}

\begin{document}

\maketitle

\begin{abstract}
  We address whether neural models for Natural Language Inference (NLI) can learn the compositional interactions between lexical entailment and negation, using four methods: the \textit{behavioral} evaluation methods of (1) challenge test sets and (2) systematic generalization tasks, and the \textit{structural} evaluation methods of (3) probes and (4) interventions. To facilitate this holistic evaluation, we present Monotonicity NLI (\MoNLI), a new naturalistic dataset focused on lexical entailment and negation. In our behavioral evaluations, we find that models trained on general-purpose NLI datasets fail systematically on \MoNLI\ examples containing negation, but that \MoNLI\ fine-tuning addresses this failure. In our structural evaluations, we look for evidence that our top-performing BERT-based model has learned to implement the monotonicity algorithm behind \MoNLI. Probes yield evidence consistent with this conclusion, and our intervention experiments bolster this, showing that the causal dynamics of the model mirror the causal dynamics of this algorithm on subsets of \MoNLI. This suggests that the BERT model at least partially embeds a theory of lexical entailment and negation at an algorithmic level.
\end{abstract}

\section{Introduction}\label{sec:introduction}

Natural Language Inference (NLI) keys into fundamental aspects of how people reason with language. Although NLI is generally cast in informal terms that embrace the indeterminacy of such reasoning, the task nonetheless manifests a number of very predictable reasoning patterns. For example, systematic manipulations of the lexical meanings \citep{glockner-etal-2018-breaking}, syntactic constructions \citep{nie2019analyzing}, and contextual assumptions \citep{pavlick-callison-burch-2016-babies} have systematic effects on the correct labels. These patterns present crisp, motivated learning targets that we can leverage to not only evaluate the ability of NLI models to learn robust solutions, but also to analyze the internal dynamics of successful models.

In this paper, our learning target concerns the role of \tech{monotonicity} in NLI \citep{MacCartney09,Icard:Moss:2013:LILT}. Specifically, we would like to determine whether models can learn to represent lexical relations and accurately model that negation reverses entailment relations (e.g., \word{dance} entails \word{move}, but \word{not move} entails \word{not dance}). This property of negation is \tech{downward monotonicity}.

In service of pursuing this question, we present Monotonicity NLI\footnote{ \url{https://github.com/atticusg/MoNLI}} (\MoNLI), a new naturalistic NLI dataset for training and assessing systems on these semantic notions (\secref{sec:datasets}). \MoNLI\ extends SNLI \citep{bowman-etal-2015-large} to provide comprehensive coverage of examples that depend on lexical reasoning with and without negation. Using \MoNLI, we conduct both behavioral and structural evaluations, seeking to provide a detailed picture of the solutions that top-performing models learn. We evaluate Enhanced Sequential Inference Models \citep{Chen-etal-2016} and BERT-based models \citep{devlin-etal-2019-bert}, along with standard baselines.

Previous work evaluating the ability of neural models to learn monotonicity has focused on challenge test sets and systematic generalization tasks \citep{yanaka-etal-2019-help, yanaka-etal-2019-neural,geiger-etal-2019-posing, richardson2019probing}. These behavioral evaluations ask whether models achieve a desired input--output behavior. We employ these methods as well, but we also ask whether models achieve an \tech{algorithmic-level} learning target, in the terms of \citet{Marr:1982:VCI:1095712}. Monotonicity reasoning can be cast as an algorithm that solves \MoNLI\ perfectly. Do neural models implement this algorithm?

We first report on two behavioral evaluations (\secref{sec:behave}). When \MoNLI\ is used as a challenge test set, we find that models trained on SNLI and/or MNLI \citep{williams:2018} fail to reason with lexical entailments when negation is involved. However, we trace these failures to gaps in the training data. In response, we pose a systematic generalization task in which we expose models to \MoNLI\ examples through fine-tuning while still requiring them to generalize to entirely new pairs of lexical items in negated linguistic contexts at test time. All our models solve the task, which suggests that they have learned general theories of lexical entailment and negation.

We then report on structural evaluations (\secref{sec:structure}), seeking to determine whether our top-performing BERT-based models implement the target monotonicity algorithm. In probing experiments, we find evidence consistent with this result, but it's not conclusive, since probes alone cannot reveal a model's causal dynamics. However, our intervention experiments provide evidence that BERT does mirror the causal dynamics of the monotonicity algorithm, at least on large subsets of \MoNLI. We conclude that this model at least partially embeds a theory of lexical entailment and negation at an algorithmic level, in addition to fully achieving the correct input--output behavior on \MoNLI.

\section{Related work}\label{sec:related-work}

\paragraph{Monotonicity}

Our empirical focus is entailment and negation. This is one (highly prevalent) aspect of monotonicity reasoning, which governs many aspects of lexical and constructional meaning in natural language \citep{SanchezValencia91,vanBenthem08NATLOG}. There is an extensive literature on monotonicity logics \citep{Moss09,Icard:2012,Icard:Moss:2013:LILT,icard-etal-2017-monotonicity}. Within NLP, \citet{maccartney-manning-2008-modeling,maccartney-manning-2009-extended} apply very rich monotonicity algebras to NLI problems, \citet{hu-etal-2019-natural, Hu2019MonaLogAL} create NLI models that use polarity-marked parse trees, and \citet{yanaka-etal-2019-neural,yanaka-etal-2019-help} and \citet{geiger-etal-2019-posing} investigate the ability of neural models to understand natural logic reasoning. While we consider only a small fragment of these approaches, the methods we develop should apply to more complex systems as well.

\paragraph{Challenge Test Sets}

Challenge\footnote{Though \textit{adversarial} and \textit{challenge} are sometimes used synonymously, we opt for the term \textit{challenge}, because our dataset was designed with the intention of evaluating whether a model learned a particular phenomenon, as opposed to breaking any particular model (cf.~\citealt{nie2019adversarial}).} test sets are supplementary evaluation resources that test the ability of a model to generalize to examples outside the distribution of the data it was trained, developed, and (standardly) tested on. These tests probe the generalization capabilities of state-of-the-art models with respect to the tasks they have been trained on, by focusing on difficult or underrepresented examples in a model's training set \citep{jia-and-liang-2017, naik-etal-2018-stress,glockner-etal-2018-breaking,richardson2019probing,talmor2019olmpics}.



\paragraph{Systematic Generalization Tasks}

\citet{Fodor:Pylyshyn:1988} offer \tech{systematicity} as a hallmark of human cognition. Systematicity says that certain behaviors are intrinsically connected to others by compositional structures. For example, understanding \word{the puppy loves Sandy} is intrinsically connected to understanding \word{Sandy loves the puppy}. For \citeauthor{Fodor:Pylyshyn:1988}, these observations trace to the mind's ability to recombine known parts and rules. There are often strong intuitions that certain generalization tasks are only solved by models with systematic structures. These tasks are referred to as \tech{systematic generalization tasks} \citep{lakeandbaroni2018,hupkes2019compositionality,yanaka2020neural, Bahdanau:2018, geiger-etal-2019-posing,goodwin2020probing}.

\paragraph{Probing}

Probes are supervised learning models trained to extract information from representations created by another model. They are a primary tool in the analysis of neural network models (\citealt{peters-etal-2018-dissecting,tenney-etal-2019-bert,clark-etal-2019-bert}; for a full review, see \citealt{belinkov-glass-2019-analysis}). In aggregate, this work has provided nuanced insights into the internal representations of these models, as well as their capacity to directly support learning diverse NLP tasks via fine-tuning \citep{hewitt-liang-2019-designing}. However, probes are only able to reveal how representations correlate with information. They cannot determine if that information plays a causal role in model predictions \citep{belinkov-glass-2019-analysis,vig2020causal}.

\paragraph{Interventions}

Intervention studies go beyond probing to make changes to the internal states of a network, with the goal of observing how those changes affect system outputs. \citet{giulianelli-etal-2018-hood} use probing results to make informed interventions during LSTM language model predictions to preserve information about the grammatical subject's number, and this led to improved performance in subject--verb agreement. \citet{vig2020causal} use interventions to characterize how gender bias is represented in the internal causal structure of a model, and find that a small number of synergistic neurons mediate gender bias. They also find that the effect of these neurons is roughly linearly separable from the effect of the remainder of the model, a remarkable finding considering the highly non-linear nature of neural networks.

\section{Monotonicity NLI dataset}\label{sec:datasets}

We created the \MoNLI\ corpus to investigate the ability of NLI models to learn the compositional interactions between lexical entailment and negation. \MoNLI\ contains 2,678 NLI examples in the usual format for NLI datasets like SNLI. In each example, the hypothesis is the result of substituting a single word $w_{p}$ in the premise for a hypernym or hyponym $w_{h}$. We refer to $w_h$ and $w_p$ as the \tech{substituted words} in an example. In 1,202 of these examples, the substitution is performed under the scope of the downward monotone operator \word{not}. Downward monotone operators reverse entailment relations: \word{dance} entails \word{move}, but \word{not move} entails \word{not dance}.  We refer to these examples collectively as \NMoNLI. In the remaining 1,476 examples, this substitution is performed under the scope of no downward monotone operator. We refer to these examples collectively as \PMoNLI.

\MoNLI\ was generated according to the following procedure. First, randomly select a premise or hypothesis sentence $s$ from the SNLI training dataset. Second, select a noun in $s$, and, using WordNet \citep{WordNet98}, select all hypernyms and hyponyms of the noun subject to two conditions: (1) the hypernym or hyponym appears in the SNLI training data, and (2) substituting the hypernym or hyponym results in a grammatical, coherent sentence $s'$. Finally, for each substitution, generate two examples for the corpus -- one where the original sentence is the premise and the edited sentence is the hypothesis, and one example with those roles reversed. Each of these example pairs has one example with the label \textbf{entailment} and one example with the label \textbf{neutral}, resulting in a dataset perfectly balanced between the two labels.

For example, suppose we select the SNLI sentence (A) and we identify the noun \word{plants} for substitution. Then we enter \word{plants} into WordNet and find that \word{flowers} is a hyponym of \word{plants}, so we substitute \word{flowers} for \word{plants} to create the edited sentence (B):
\begin{center}
\setlength{\tabcolsep}{2pt}
\begin{tabular}{r l@{ \ }c}
(A) & The three children are not holding & \textbf{plants}. \\
    & & $\Downarrow$ \\
(B) &  The three children are not holding & \textbf{flowers}.
\end{tabular}
\end{center}
This  leads to two new \MoNLI\ examples:
\begin{center}
\begin{tabular}{r c l}
(A) & \textbf{entailment} & (B) \\
(B) & \textbf{neutral}    & (A)
\end{tabular}
\end{center}

These two examples would belong to NMoNLI, due to \textit{not} scoping over the substitution site. If \textit{not} were removed from both of these sentences, then their labels would be swapped and both examples would belong to PMoNLI.

MoNLI was generated by the authors by hand; examples judged to be unnatural were removed, and any grammatical or spelling errors in the original SNLI sentence were corrected.

This data generation process is similar to that of \citet{glockner-etal-2018-breaking}, except they focus on the lexical relations of exclusion and synonymy, while we focus on entailment relations. This difference prevents their dataset from capturing monotonicity reasoning, which involves entailment relations, but not exclusion or synonymy.

\begin{table*}[tp]
  \centering
  \small
  \setlength{\tabcolsep}{4pt}
    \begin{tabular}{l c c@{\hspace{40pt}} c c c @{\hspace{40pt}} c c}
      \toprule
      \multicolumn{3}{c@{\hspace{40pt}}}{} & \multicolumn{3}{c@{\hspace{40pt}}}{\textbf{No \MoNLI\ fine-tuning}} & \multicolumn{2}{@{\hspace{-14pt}}c}{\textbf{With \NMoNLI\ fine-tuning}} \\
      Model  & Input pretraining & NLI train data & SNLI & PMoNLI & NMoNLI  & SNLI & NMoNLI \\
      \midrule
      CBOW & GloVe        & SNLI train    &    78.9       & 64.6   & 22.9 & 65.9 & 95.5      \\
      BiLSTM   & GloVe        & SNLI train    &  81.6   &  73.2 &  37.9  & 74.6 & 93.5     \\
      ESIM   & GloVe        & SNLI train    & 87.9       & 86.6 & 39.4  &56.9 & 96.2\\
      ESIM   & GloVe        & & --       &--  &-- & -- & 98.0 \\
      ESIM   & & &    --    &--  &--   & -- & 35.5 \\
      BERT   & BERT         & SNLI train    & 90.8       & 94.4 & \phantom{0}2.2  & 90.5 & 90.0 \\
      BERT   & BERT         &               & --          & --    & --  &  --    & 96.7 \\
      BERT   &              &               & --         & --   & --    & -- & 62.3 \\
    \bottomrule
    \end{tabular}
    \caption{The results of our behavioral analysis. The columns labeled \textit{No \MoNLI\ fine-tuning} display the challenge test set results (\secref{sec:challenge}), and the columns labeled \textit{With \MoNLI\ fine-tuning} display systematic generalization task results (\secref{sec:gen:fine-tune}).  The numbers are accuracy values; all the datasets have balanced label distributions. Dashes mark experiments that would involve untrained NLI parameters due to training/fine-tuning set-up.}
    \label{tab:adv}
\end{table*}

\section{Models}

We evaluated four models on MoNLI:
\begin{description}\setlength{\itemsep}{0pt}
\item[CBOW] The continuous bag of words baseline from \citet{williams:2018}.

\item[BiLSTM] The bidirectional LSTM baseline from \citet{williams:2018}.


\item[ESIM] The Enhanced Sequential Inference Model \citep{Chen-etal-2016} is a hybrid TreeLSTM-based and biLSTM-based model that uses an inter-sentence attention mechanism to align words across sentences.


\item[BERT] A Transformer model trained to do masked language modeling and next-sentence prediction \citep{devlin-etal-2019-bert}. We rely on uncased BERT-base parameters from Hugging Face \texttt{transformers} \citep{Wolf2019HuggingFacesTS}.
\end{description}

The first two models serve as baselines, while the other two models achieve comparable, near state-of-the-art scores on SNLI.


\section{Behavioral Evaluations}\label{sec:behave}


\subsection{\MoNLI\ as a Challenge Test Set}\label{sec:challenge}

We first use \MoNLI\ as a challenge test dataset, i.e., models trained only on SNLI are expected to generalize to \MoNLI. \MoNLI\ can be considered a challenge test dataset that evaluates an NLI model's ability to perform simple inferences founded in lexical entailments and monotonicity. As discussed in \secref{sec:datasets}, it is not especially adversarial, in that we sampled sentences from the SNLI training set and only substituted in hypernyms and hyponyms that occur in the SNLI training set. This keeps \MoNLI\ as close as possible to the distribution of SNLI. Thus, if a model fails on \MoNLI, we can be confident that this failure stems from a lack of knowledge about monotonicity and lexical entailment relations, rather than some other confounding factor like syntactic structures or vocabulary items that were unseen in training.

\subsubsection{Results}

The results are in \tabref{tab:adv} under the heading `No \MoNLI\ fine-tuning', and they are stark. The four models achieve comparably high accuracies on SNLI and PMoNLI, the examples where no downward monotone operators scope over the substitution site. However, they are well below chance accuracy on NMoNLI, the examples where \word{not} scopes over the substitution site. BERT is more extreme than the other models, achieving a higher accuracy on PMoNLI than SNLI and almost zero accuracy on NMoNLI. High performance on PMoNLI shows that models have knowledge of the lexical relations between the substituted words, but low performance on NMoNLI shows the models have no knowledge of the downward monotone nature of \word{not}. In fact, the below chance accuracy on NMoNLI indicates that these models are somewhat reliably (incredibly reliably in BERT's case) predicting the wrong label on these examples, suggesting that they treat NMoNLI examples the same as PMoNLI examples.

\subsubsection{Discussion}

While these models trained on SNLI do not know that \word{not} is downward monotone in these examples, this is not conclusive evidence that they are unable to learn this semantic property. This ability might not be necessary for success on SNLI, where only 38 examples have negation in both the premise and hypothesis. A natural next step is to train on MNLI, where the coverage with regard to negation is better: about 18K examples (${\approx}$4\%) have negation in the premise and hypothesis. We tried this, by combining MNLI with SNLI, and the results were almost exactly the same. However, even the MNLI examples might not manifest the kind of monotonicity reasoning that we are targeting. Our next experiments help to resolve this issue.

\subsection{A Systematic Generalization Task }\label{sec:gen:fine-tune}

Our three models trained on SNLI have knowledge of the lexical relations between substituted words, but do not know that the presence of \word{not} reverses the relationship between the word-level relation and the sentence-level relation. We now conduct a behavioral evaluation to determine whether models are able to learn a general theory of lexical entailment and negation when exposed to a limited subset of NMoNLI during training.

In designing systematic generalization tasks, we seek to constrain the training data in ways that prevent unsystematic models from succeeding. Defining disjoint train/test splits is enough to foil truly unsystematic models (e.g., simple look-up tables). However, building on much previous work \citep{lakeandbaroni2018,hupkes2019compositionality,yanaka2020neural, Bahdanau:2018,goodwin2020probing,geiger-etal-2019-posing}, we contend that a randomly constructed disjoint train/test split only diagnoses the most basic level of systematicity. More difficult systematic generalization tasks will only be solved by models exhibiting more complex compositional structures. Specifically, we want our systematic generalization task to be solved only by models that compute lexical entailment relations that may be reversed by negation. A learning model that memorizes labels based on substituted word pairs and whether negation is present would succeed on a disjoint train and test set as long as all pairs of substituted words appear during training, and this model does not compute the lexical relation between word pairs.

As such, we propose a generalization task where \NMoNLI\ is partitioned into train and test sets such that the substituted words in the train set and the substituted words in the test sets are disjoint.\footnote{We use only \NMoNLI\ in our systematic generalization task because models trained on SNLI already achieve high performance on \PMoNLI.} The specific train/test split we used is described in \appref{app:split}. Ideally, a model trained on SNLI that is further trained on NMoNLI will still maintain strong performance on SNLI.  We use inoculation by fine-tuning \citep{liu-etal-2019-inoculation} to evaluate models on this ability. We report on the inoculated model with the highest average performance on SNLI test and \NMoNLI\ test (full details of the inoculation process are in \appref{app:inoc}).

The models are evaluated on examples where they know the relation between the substituted words, as evidenced by high performance on PMoNLI, but have not seen those substituted words in the presence of negation during training. However, they have seen other substituted words with the same relation in the presence of negation during training, making this task \textit{hard}, but \textit{fair} \citep{geiger-etal-2019-posing}. To solve this harder generalization task, we believe a model must learn to reverse the lexical relation \emph{in general}; the identity of the substituted words must be abstracted away.

\subsubsection{Results and Discussion}\label{sec:gen:results}

We present our results in \tabref{tab:adv}, under the heading `With \NMoNLI\ fine-tuning'. All of our models solve this generalization task. However, only BERT does so while maintaining high performance on SNLI. We also report ablation studies on our two non-baseline models, evaluating their performance on our systematic generalization task without training on SNLI and without any pretraining at all. We find that both models still succeed with no pretraining on SNLI, but fail with no pretraining whatsoever. This suggests that BERT pretraining and GloVe vectors both provide sufficient information about lexical relations for the models to succeed. BERT's ability to get slightly above chance performance with no pretraining indicates the presence of some statistical artifacts in our dataset \citep{gururangan-etal-2018-annotation}.



In sum, our models were able to solve our systematic generalization task, which we believe to be evidence that they learn to compute the lexical relations between substituted words. However, we also believe this evidence is weak, as there is no formal relationship between a model solving a generalization task and that model having any particular systematic internal structures. This evaluation is fundamentally behavioral, only concerning model inputs and outputs. We believe that a structural evaluation is necessary to conclusively evaluate systematicity.

\begin{figure}[!tp]
\input{algorithmfigure}
\caption{An algorithm able to solve the \MoNLI\  dataset that provides a theoretically motivated learning target for neural models at an algorithmic level of analysis  \citep{Marr:1982:VCI:1095712}. $\proc{Infer}$ takes in an example from \MoNLI\ and outputs the relation between the premise and hypothesis.  It uses three predefined functions. \lexrelproc\ returns the relation (one of $\entailrelations$) between the substituted words in the premise and hypothesis. \proc{contains-not} returns true iff negation is present. \proc{reverse} maps $\sqsubset$ to $\sqsupset$ and vice-versa.}
\label{alg:1}
\end{figure}

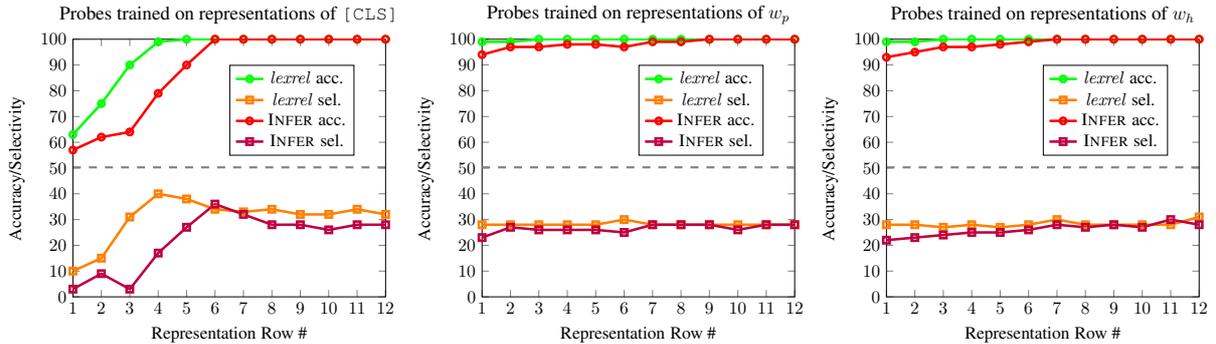
\begin{figure*}[tp]
  \centering
  \begin{tikzpicture}[scale=0.6, line width=2mm]

\draw [thick,dashed,color=black!50] (0,2.86) -- (7,2.86);
\begin{axis}[
    title={\large Probes trained on representations of ${\CLS}$},
    every axis plot/.append style={ultra thick},
    xlabel={Representation Row \#},
    ylabel={Accuracy/Selectivity},
    xmin=1, xmax=12,
    ymin=0, ymax=100,
    ytick={0, 10,20,30,40,50,60,70,80,90,100},
    xtick={1,2,3,4,5,6,7,8,9,10,11,12},
    legend style={at={(0.9, 0.9)}},
    mark size=2pt
]

\addplot[
    color=green,
    mark=o,
    ]
    coordinates {
    (1,63)(2,75)(3,90)(4,99)(5,100)(6,100)(7,100)(8,100)(9,100)(10,100)(11,100)(12,100)
    };
\addlegendentry{\lexvar\ acc.}

\addplot[
    color=orange,
    mark=square,
    ]
    coordinates {
    (1,10)(2,15)(3,31)(4,40)(5,38)(6,34)(7,33)(8,34)(9,32)(10,32)(11,34)(12,32)
    };
\addlegendentry{\lexvar\ sel.}

\addplot[
    color=red,
    mark=o,
    ]
    coordinates {
    (1,57)(2,62)(3,64)(4,79)(5,90)(6,100)(7,100)(8,100)(9,100)(10,100)(11,100)(12,100)
    };
\addlegendentry{$\proc{Infer}$ acc.}

\addplot[
    color=purple,
    mark=square,
    ]
    coordinates {
    (1,3)(2,9)(3,3)(4,17)(5,27)(6,36)(7,32)(8,28)(9,28)(10,26)(11,28)(12,28)
    };
\addlegendentry{$\proc{Infer}$ sel.}
    
\end{axis}

\end{tikzpicture}
\begin{tikzpicture}[scale=0.6, line width=2mm]

\begin{axis}[
    title={\large Probes trained on representations of ${w_p}$},
    every axis plot/.append style={ultra thick},
    xlabel={Representation Row \#},
    ylabel={Accuracy/Selectivity},
    xmin=1, xmax=12,
    ymin=0, ymax=100,
    ytick={0, 10,20,30,40,50,60,70,80,90,100},
    xtick={1,2,3,4,5,6,7,8,9,10,11,12},
    legend style={at={(0.9, 0.9)}},
    mark size=2pt
]

\addplot[
    color=green,
    mark=o,
    ]
    coordinates {
    (1,99)(2,99)(3,100)(4,100)(5,100)(6,100)(7,100)(8,100)(9,100)(10,100)(11,100)(12,100)
    };
\addlegendentry{\lexvar\ acc.}
\addplot[
    color=orange,
    mark=square,
    ]
    coordinates {
    (1,28)(2,28)(3,28)(4,28)(5,28)(6,30)(7,28)(8,28)(9,28)(10,28)(11,28)(12,28)
    };
\addlegendentry{\lexvar\ sel.}

\addplot[
    color=red,
    mark=o,
    ]
    coordinates {
    (1,94)(2,97)(3,97)(4,98)(5,98)(6,97)(7,99)(8,99)(9,100)(10,100)(11,100)(12,100)
    };
\addlegendentry{$\proc{Infer}$ acc.}

\addplot[
    color=purple,
    mark=square,
    ]
    coordinates {
    (1,23)(2,27)(3,26)(4,26)(5,26)(6,25)(7,28)(8,28)(9,28)(10,26)(11,28)(12,28)
    };
\addlegendentry{$\proc{Infer}$ sel.}
    
\end{axis}

\draw [thick,dashed,color=black!50] (0,2.86) -- (7,2.86);
\end{tikzpicture}
\begin{tikzpicture}[scale=0.6, line width=2mm]

\draw [thick,dashed,color=black!50] (0,2.86) -- (7,2.86);
\begin{axis}[
    title={\large Probes trained on representations of ${w_h}$},
    every axis plot/.append style={ultra thick},
    xlabel={Representation Row \#},
    ylabel={Accuracy/Selectivity},
    xmin=1, xmax=12,
    ymin=0, ymax=100,
    ytick={0, 10,20,30,40,50,60,70,80,90,100},
    xtick={1,2,3,4,5,6,7,8,9,10,11,12},
    legend style={at={(0.9, 0.9)}},
    mark size=2pt
]
\addplot[
    color=green,
    mark=o,
    ]
    coordinates {
    (1,99)(2,99)(3,100)(4,100)(5,100)(6,100)(7,100)(8,100)(9,100)(10,100)(11,100)(12,100)
    };
\addlegendentry{\lexvar\ acc.}

\addplot[
    color=orange,
    mark=square,
    ]
    coordinates {
    (1,28)(2,28)(3,27)(4,28)(5,27)(6,28)(7,30)(8,28)(9,28)(10,28)(11,28)(12,31)
    };
\addlegendentry{\lexvar\ sel.}

\addplot[
    color=red,
    mark=o,
    ]
    coordinates {
    (1,93)(2,95)(3,97)(4,97)(5,98)(6,99)(7,100)(8,100)(9,100)(10,100)(11,100)(12,100)
    };
\addlegendentry{$\proc{Infer}$ acc.}

\addplot[
    color=purple,
    mark=square,
    ]
    coordinates {
    (1,22)(2,23)(3,24)(4,25)(5,25)(6,26)(7,28)(8,27)(9,28)(10,27)(11,30)(12,28)
    };
\addlegendentry{$\proc{Infer}$ sel.}
    
\end{axis}

\end{tikzpicture}


  \caption{Results where classifier probes are trained on BERT representations to predict the value of \lexvar\ and the output of $\proc{Infer}$ (\figref{alg:1}). The grey dotted line provides a soft ceiling for selectivity values, because we expect control probes trained on a binary task to at least achieve chance accuracy. }
  \label{fig:probe}
\end{figure*}

\section{Structural Evaluations}\label{sec:structure}

In our behavioral evaluations, the learning target was to mimic the input--output behavior defined by \MoNLI. Assessing this learning target is straightforward. We now report on structural evaluations to try to determine whether a neural model has particular internal dynamics. For this, we rely on very recent probing and intervention methodologies that are not yet well understood and must be tailored to the model being analyzed. As such, we choose to focus on a single model, namely, the BERT model from \secref{sec:behave} fine-tuned on NMoNLI. We chose BERT because it achieved exceptional results on \NMoNLI\ after fine-tuning without experiencing a significant drop on SNLI.

\Figref{alg:1} presents the simple algorithm $\proc{Infer}$, which is our learning target. It takes in a \MoNLI\ example and stores the lexical entailment relation between the substituted words in the variable \lexvar. If negation is present, the reverse of \lexvar\ is returned; if there is no negation, \lexvar\ itself is returned. This is simply an algorithmic description of the \MoNLI\ construction method. The most important piece is the intermediate variable \lexvar. Intuitively, if our BERT model implements this algorithm, there will be some representation in BERT that \textit{stores} \lexvar\ and BERT will \textit{use} that representation for a final prediction. Probes can give us an idea of where information is stored, and interventions help us see how that information is used.

Before we can go looking for where BERT stores and uses \lexvar, we must limit ourselves to a tractable number of model internal representations. When our BERT model processes an example from \MoNLI, it is tokenized as
\begin{center}
  $e = \langle \CLS, p, \SEP, h, \SEP \rangle$
\end{center}
and 12 rows of vector representations are created, so each token is associated with 12 vectors.
We localize our efforts to the representations created for $\CLS$ and the tokens for the substituted words in the premise and hypothesis, $w_p$ and $w_h$ (as described in \secref{sec:datasets}). This narrows our search to 36 possible vector locations where BERT could be storing the variable \lexvar\ for use in final output prediction. We denote these 36 locations with  $\BERT_{w_p}^{r}$, $\BERT_{w_h}^{r}$, and $\BERT_{\CLS}^{r}$ where $r$ is a row ($1 \leqslant r \leqslant 12$).

\subsection{Probes}\label{sec:probe-compare}

We follow \citet{hupkes-etal-2018-analysing} in using probing evidence to determine whether a neural model stores the same information as a symbolic algorithm. They used probes to predict variable values used in an algorithm from the hidden states of sequential recurrent networks trained to perform basic arithmetic. We do something similar, probing the 36 vector locations defined by $\BERT_{w_p}^{r}$, $\BERT_{w_h}^{r}$, and $\BERT_{\CLS}^{r}$ for the value of the variable \lexvar\ and the output of $\proc{Infer}$.

\citet{hewitt-liang-2019-designing} argue that accuracy is a poor metric for probes and that the ideal probe will highly \tech{selective}, that is, it will have high accuracy on a linguistic task but low accuracy on a control task where inputs are given random labels. In this setting, our linguistic tasks are predicting the value of $\lexvar$ and the output of $\proc{Infer}$ from a model-internal vector created by BERT for some \MoNLI\ example. Our control task is identical, except labels are randomly assigned to inputs. \citeauthor{hewitt-liang-2019-designing} demonstrate that small, linear probes result in high selectivity. Following this guidance, we used a linear classifier with 4 hidden units that was trained and evaluated on all of \MoNLI.


Our probing results are summarized in \figref{fig:probe}. Probes were able to achieve high accuracy and high selectivity predicting the output of $\proc{Infer}$ at every location other than the locations $\BERT_{\CLS}^k$ where $1 \leq k\leq 4$, and high accuracy and high selectivity predicting the value of $\lexvar$ at every location other than $\BERT_{\CLS}^1$ and $\BERT_{\CLS}^2$.

This qualitative picture is compatible with a story where BERT stores the value of \lexvar\ at any location other than $\BERT_{\CLS}^1$ or $\BERT_{\CLS}^2$ and then uses this information to compute a final output prediction at any location other than the locations $\BERT_{\CLS}^k$ where $1 \leq k\leq 4$. The fact that probes trained on the vectors at locations $\BERT_{\CLS}^3$ or $\BERT_{\CLS}^4$ have high accuracy and selectivity predicting the value of \lexvar, but moderate accuracy and low selectivity predicting the output of $\proc{Infer}$ may suggest a more specific story where these two locations store the value of the variable \lexvar\ before this information is used to compute the final output.

We emphasize that, while the probing results are compatible with these stories, they only provide conclusive evidence about how representations correlate with the value of \lexvar\ and the output of $\proc{Infer}$. They cannot determine whether this information plays a causal role in model predictions \citep{belinkov-glass-2019-analysis,vig2020causal}.

\subsection{Interventions}\label{sec:interv}

Probes give us a picture of where information is stored by our BERT model, but they cannot determine whether that information is used to make final predictions. Interventions can help us address this deeper question. As discussed above, our algorithmic-level learning target is for BERT to mimic the dynamics of the algorithm $\proc{Infer}$ in \figref{alg:1}. \citet{icard2017} provided the insight that algorithms like $\proc{Infer}$ can be explicitly understood as causal models \citep{pearl}. This means that the causal role of $\lexvar$, the lone variable in \proc{Infer}, can be characterized with counterfactual claims about how altering the value of the variable would cause output behavior to change.

Suppose \proc{Infer} is run on a \MoNLI\ example $i$. Let $\varfunc(i) \in \entailrelations$ be the value that $\lexvar$ takes on, and let $\proc{infer}(i) \in \entailrelations$ be the output. Then \proc{Infer} can be see as providing the following counterfactual characterization of $\lexvar$: if the value of $\lexvar$ were changed from $\varfunc(i)$ to $\varfunc(j)$, where $j$ is a second \MoNLI\ example, then $\proc{infer}(i)$ would change to
\begin{multline*}
\interfunc{i}{j}{\varfunc} = \\
\begin{cases}
\outfunc(i) & { \varfunc(i) = \varfunc(j)} \\
 \proc{reverse}(\outfunc(i)) & {\varfunc(i) \not = \varfunc(j)} \\
 \end{cases}
\end{multline*}
In other words, if $\lexvar$ were to take on the opposite value, then the output would also take on the opposite value.

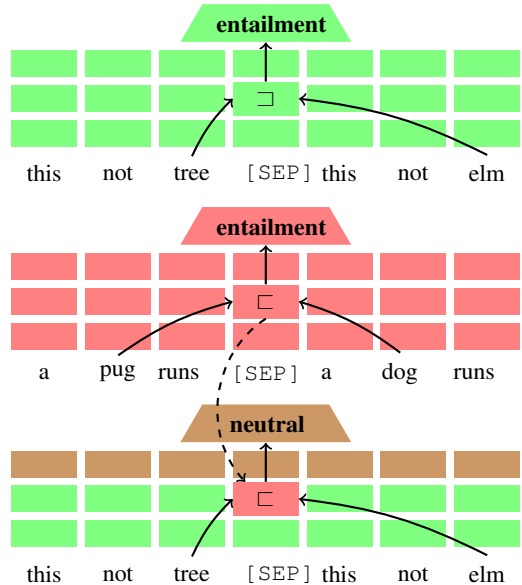
\begin{figure}
\centering
\newcommand{\repgap}{0.1cm}

\begin{tikzpicture}
  \tikzset{
    every node=[
    draw,
    font=\footnotesize,
    shape=rectangle,
    align=center,
    minimum width=0.6cm,
    minimum height=0.35cm,
    text width=0.6cm
    ],
    input/.style={
    },
    hiddenRed/.style={
      fill=green!50
    },
    hiddenBlue/.style={
      fill=red!50
    },
    outputBlue/.style={
      trapezium,
      fill=red!50,
      text=black,
      minimum height=0.5cm,
      minimum width=0.75cm
    },
    hiddenPurple/.style={
      fill=brown!80
    },
    outputPurple/.style={
      trapezium,
      fill=brown!80,
      text=black,
      minimum height=0.5cm,
      minimum width=0.75cm
    },
    outputRed/.style={
      trapezium,
      fill=green!50,
      text=black,
      minimum height=0.5cm,
      minimum width=0.75cm
    },
    outputLabel/.style={
    },
    swap/.style={
      solid,
      thick,
      ->,
      shorten <=0pt,
      shorten >=0pt
    },
    replace/.style={
      dashed,
      thick,
      ->,
      shorten <=0pt,
      shorten >=0pt
    },
    ann/.style={
      minimum width=0.7cm,
      minimum height=0.5cm,
      text width=5cm,
      align=left,
    }   
};


  \node[input](x1){a};
  \node[input, right=\repgap of x1](x2){pug};
  \node[input, right=\repgap of x2,xshift=-1ex](x3){runs};
  \node[input, right=\repgap of x3](x4){\SEP};
  \node[input, right=\repgap of x4](x5){a};
  \node[input, right=\repgap of x5](x6){dog};
  \node[input, right=\repgap of x6](x7){runs};

  \node[hiddenBlue, above=\repgap of x1](h11){};
  \node[hiddenBlue, right=\repgap of h11](h12){};
  \node[hiddenBlue, right=\repgap of h12](h13){};
  \node[hiddenBlue, right=\repgap of h13](h14){};
  \node[hiddenBlue, right=\repgap of h14](h15){};
  \node[hiddenBlue, right=\repgap of h15](h16){};
  \node[hiddenBlue, right=\repgap of h16](h17){};

  \node[hiddenBlue, above=\repgap of h11](h21){};
  \node[hiddenBlue, right=\repgap of h21](h22){};
  \node[hiddenBlue, right=\repgap of h22](h23){};
  \node[hiddenBlue, right=\repgap of h23](h24){$\sqsubset$};
  \node[hiddenBlue, right=\repgap of h24](h25){};
  \node[hiddenBlue, right=\repgap of h25](h26){};
  \node[hiddenBlue, right=\repgap of h26](h27){};

  \node[hiddenBlue, above=\repgap of h21](h31){};
  \node[hiddenBlue, right=\repgap of h31](h32){};
  \node[hiddenBlue, right=\repgap of h32](h33){};
  \node[hiddenBlue, right=\repgap of h33](h34){};
  \node[hiddenBlue, right=\repgap of h34](h35){};
  \node[hiddenBlue, right=\repgap of h35](h36){};
  \node[hiddenBlue, right=\repgap of h36](h37){};

  \node[outputBlue, above=\repgap of h34](o1){};
  \node[above=\repgap of h34,xshift=-2ex](label1){\centering \textbf{entailment}};


  \node[input, below=2.2cm of x1](xx1){this};
  \node[input, right= \repgap of xx1](xx2){not};
  \node[input, right= \repgap of xx2](xx3){tree};
  \node[input, right=\repgap of xx3](xx4){$\SEP$};
  \node[input, right=\repgap of xx4](xx5){this};
  \node[input, right=\repgap of xx5](xx6){not};
  \node[input, right=\repgap of xx6](xx7){elm};

  \node[hiddenRed, above=\repgap of xx1](hh11){};
  \node[hiddenRed, right=\repgap of hh11](hh12){};
  \node[hiddenRed, right=\repgap of hh12](hh13){};
  \node[hiddenRed, right=\repgap of hh13](hh14){};
  \node[hiddenRed, right=\repgap of hh14](hh15){};
  \node[hiddenRed, right=\repgap of hh15](hh16){};
  \node[hiddenRed, right=\repgap of hh16](hh17){};

  \node[hiddenRed, above=\repgap of hh11](hh21){};
  \node[hiddenRed, right=\repgap of hh21](hh22){};
  \node[hiddenRed, right=\repgap of hh22](hh23){};
  \node[hiddenBlue, right=\repgap of hh23](hh24){$\sqsubset$};
  \node[hiddenRed, right=\repgap of hh24](hh25){};
  \node[hiddenRed, right=\repgap of hh25](hh26){};
  \node[hiddenRed, right=\repgap of hh26](hh27){};

  \node[hiddenPurple, above=\repgap of hh21](hh31){};
  \node[hiddenPurple, right=\repgap of hh31](hh32){};
  \node[hiddenPurple, right=\repgap of hh32](hh33){};
  \node[hiddenPurple, right=\repgap of hh33](hh34){};
  \node[hiddenPurple, right=\repgap of hh34](hh35){};
  \node[hiddenPurple, right=\repgap of hh35](hh36){};
  \node[hiddenPurple, right=\repgap of hh36](hh37){};

  \node[outputPurple, above=\repgap of hh34](o2){};
  \node[above=\repgap of hh34,xshift=-1ex](label2){\textbf{neutral}};

    \node[input, above=2.2cm of x1](xxx1){this};
  \node[input, right= \repgap of xxx1](xxx2){not};
  \node[input, right= \repgap of xxx2](xxx3){tree};
  \node[input, right=\repgap of xxx3](xxx4){$\SEP$};
  \node[input, right=\repgap of xxx4](xxx5){this};
  \node[input, right=\repgap of xxx5](xxx6){not};
  \node[input, right=\repgap of xxx6](xxx7){elm};

  \node[hiddenRed, above=\repgap of xxx1](hhh11){};
  \node[hiddenRed, right=\repgap of hhh11](hhh12){};
  \node[hiddenRed, right=\repgap of hhh12](hhh13){};
  \node[hiddenRed, right=\repgap of hhh13](hhh14){};
  \node[hiddenRed, right=\repgap of hhh14](hhh15){};
  \node[hiddenRed, right=\repgap of hhh15](hhh16){};
  \node[hiddenRed, right=\repgap of hhh16](hhh17){};

  \node[hiddenRed, above=\repgap of hhh11](hhh21){};
  \node[hiddenRed, right=\repgap of hhh21](hhh22){};
  \node[hiddenRed, right=\repgap of hhh22](hhh23){};
  \node[hiddenRed, right=\repgap of hhh23](hhh24){$\sqsupset$};
  \node[hiddenRed, right=\repgap of hhh24](hhh25){};
  \node[hiddenRed, right=\repgap of hhh25](hhh26){};
  \node[hiddenRed, right=\repgap of hhh26](hhh27){};

  \node[hiddenRed, above=\repgap of hhh21](hhh31){};
  \node[hiddenRed, right=\repgap of hhh31](hhh32){};
  \node[hiddenRed, right=\repgap of hhh32](hhh33){};
  \node[hiddenRed, right=\repgap of hhh33](hhh34){};
  \node[hiddenRed, right=\repgap of hhh34](hhh35){};
  \node[hiddenRed, right=\repgap of hhh35](hhh36){};
  \node[hiddenRed, right=\repgap of hhh36](hhh37){};

  \node[outputRed, above=\repgap of hhh34](o3){};
  \node[above=\repgap of hhh34,xshift=-2ex](label2){\textbf{entailment}};

  \path(h24.south) edge[replace, bend right=50pt] (hh24);

  \path(xxx3.north) edge[swap, bend left=10pt] (hhh24.west);
  \path(xxx7.north) edge[swap, bend right=10pt] (hhh24.east);
  \path(hhh24.north) edge[swap] (o3.south);
  \path(x2.north) edge[swap, bend left=10pt] (h24.west);
  \path(x6.north) edge[swap, bend right=10pt] (h24.east);
  \path(h24.north) edge[swap] (o1.south);
  \path(xx3.north) edge[swap, bend left=10pt] (hh24.west);
  \path(xx7.north) edge[swap, bend right=10pt] (hh24.east);
  \path(hh24.north) edge[swap] (o2.south);
\end{tikzpicture}
\vspace{-2mm}
\caption{An illustrative \textbf{interchange intervention}:  The solid arrows represent a hypothesis about where the model stores and uses information about lexical entailment. The dotted arrow is an interchange intervention, where the green vector (top) we think stores reverse entailment, trees $\sqsupset$ elms, is interchanged with the red vector (middle) we think stores forward entailment, pugs $\sqsubset$ dogs, leading to a modified network (bottom). If our hypothesis is correct, then the  output should change from \textbf{entailment} to \textbf{neutral}, because the negation in the green example reverses the relationship between lexical entailment and sentence-level entailment. If this label reversal is not observed, crucial entailment information must lie elsewhere in the network.}
\label{fig:interchange-schematic}
\end{figure}

Our analytic tool for evaluating whether such causal dynamics are present in BERT is the \tech{interchange intervention}. \Figref{fig:interchange-schematic} provides a high-level picture of how these experiments work, and the following definition seeks to make this more precise and general:


\paragraph{Interchange Intervention}
 Let $L$ be one of the 36 locations defined by $\BERT_{w_p}^{r}$, $\BERT_{w_h}^{r}$, and $\BERT_{\CLS}^{r}$. When BERT is making a prediction for $i$, suppose that the vector created at location $L$ on input $i$ is replaced with the vector created at location $L$ on input $j$ and this results in the output $y$. We say that $y$ is the result of an interchange intervention from $i$ to $j$ at location $L$ and denote this output as $\interv{i}{j}{L}$.
\\[10pt]
In essence, $\interv{i}{j}{L}$ characterizes the output behavior that results from an experiment where model-internal vectors are interchanged at location $L$. Recall that $\interfunc{i}{j}{\varfunc}$ describes what output is provided by \proc{Infer} if variables are interchanged. If for some subset of \MoNLI\ $S$, we believe that BERT is both storing the value of \lexvar\ at some location $L$ and using that information to make a final prediction, then for all $i,j \in S$ the following should hold:
\begin{align*}
\interfunc{i}{j}{\varfunc} &= \interv{i}{j}{\bertmap}
\end{align*}
This amounts to observing that the variables in the algorithm and the vectors in the model satisfy the same counterfactual claims. When a vector representing forward entailment is interchanged with a different vector representing forward entailment, model output behavior should be unchanged. If a vector representing forward entailment is interchanged with a different vector representing reverse entailment, then the model output should be reversed.

\paragraph{Results}

Due to computational constraints, we randomly conducted interchange experiments at our 36 different locations and chose the location with the most promise, namely, $\BERT^3_{w_h}$. (Appendix~\ref{app:interv} covers our selection methodology in detail.) We conducted $\approx$7 million interchange experiments at this location, one experiment for every pair of examples in \MoNLI. Using a simple greedy algorithm, we discovered several large subsets of \MoNLI\ where \BERT\ mimics the causal dynamics of $\proc{Infer}$. (The greedy algorithm is described in Appendix~\ref{app:interv}.) These subsets have size 98, 63, 47, and 37, and for each of these subsets there are many pairs of examples with interchange experiments that had a causal impact on the final model prediction. To put these results in context, if interchange experiments had a random effect on model output, then the expected number of subsets larger than 20 with this property would be less than $10^{-8}$.



\paragraph{Discussion}



These results show that the values assigned by the algorithm $\proc{Infer}$ to the variable \lexvar\ and the vectors created by BERT at the location $\BERT_{w_h}^3$ exhibit the same causal dynamics on four large subsets of \MoNLI. In Appendix~\ref{app:interv} we show a visualization of the subset with 98 examples. These pairs contain only 13 of the 69 distinct hyponyms in MoNLI, which makes it clear that this subset of \MoNLI\ is not a random sample, but rather reflects a coherent semantic space. From this we conclude that, in addition to capturing the input--output behavior described by \MoNLI, our BERT model at least partially embeds a theory of lexical entailment and negation at an algorithmic level of analysis.

Importantly, these results do not show that BERT fails to mimic the causal dynamics of $\proc{Infer}$ on larger subsets of \MoNLI. First, we only conducted interchange experiments for every pair of examples in \MoNLI\ at the location $\BERT^3_{w_h}$. Second, we did not consider the possibility that BERT stores  and uses the value of \lexvar\ at different locations, depending on which input is provided. Third, analyzing vector representations may be too coarse-grained; perhaps experiments will need to be done on individual vector units. Finally, we used a greedy algorithm to discover the four subsets of \MoNLI.
We did not exhaustively analyze BERT to find the largest subset of \MoNLI\ on which it mimics the causal dynamics of $\proc{Infer}$; such an analysis is likely computationally impossible. What we did do is perform an efficient analysis that was able to find several large subsets of \MoNLI\ on which the desired causal dynamics are present.

\section{Conclusion}\label{sec:conclusion}

To operationalize our research question of whether neural NLI models can learn the compositional interactions between lexical entailment and negation, we constructed two learning targets for neural NLI models: (1) learn the input--output behavior described by \MoNLI\ and (2) acquire the internal dynamics of the algorithm $\proc{Infer}$. We evaluated the first learning target with two behavioral evaluation methods, using challenge datasets to show that state-of-the-art models trained on general-purpose NLI datasets fail to exhibit the correct behavior when negation is present and then following up with a systematic generalization task that showed our models are able to learn the correct input--output behavior when fine-tuned on a limited, but sufficient, subset of \NMoNLI. We evaluated the second learning target with two structural evaluation methods, using probes to investigate where information about the variable \lexvar\ from $\proc{Infer}$ might be stored in a BERT model and using interventions to show that on some subsets of \MoNLI\ our BERT model exhibits the same causal dynamics as the algorithm $\proc{Infer}$.

We believe that our holistic evaluation, leveraging both behavioral and structural methods, provides a multifaceted picture of how neural NLI models treat lexical entailment and negation. While our interchange intervention methodology is not yet formally grounded, there is great promise in the idea of investigating whether a neural model mirrors the causal dynamics of an algorithm.




\bibliography{anthology,systematicity-bib}
\bibliographystyle{acl_natbib}

\newpage
\clearpage
\appendix
 \section{Appendices}
 \label{sec:appendix}

\subsection{Train--Test Split for Systematic Generalization Task}\label{app:split}

In our systematic generalization task, \NMoNLI\ is partitioned into train, dev, and test sets such that the substituted words in the train set and the substituted words in the dev and test sets are disjoint. The specific train/test split we used is described in \tabref{tab:hyponym}.

\begin{table}[h]
    \centering
    \begin{tabular}{lr @{\hspace{45pt}} lr}
\toprule
\multicolumn{2}{c@{\hspace{45pt}}}{NMoNLI Train} & \multicolumn{2}{c}{NMoNLI Test}\\
\midrule
person & 198 & dog & 88 \\
instrument & 100 & building & 64 \\
food & 94 & ball & 28 \\
machine & 60 & car & 12 \\
woman & 58 & mammal & 4 \\
music & 52 & animal & 4 \\
tree & 52 & & \\
boat & 46  & & \\
fruit & 42  & & \\
produce & 40  & & \\
fish & 40 & &  \\
plant & 38 & &  \\
jewelry & 36 & &  \\
anything & 34 & &  \\
hat & 20 & &  \\
man & 20  & & \\
horse & 16& & \\
gun & 12 & &\\
adult & 10& & \\
shirt & 8 & &\\
shoe & 6 & &\\
store & 6 & &\\
cake & 4 & &\\
individual & 4& & \\
clothe & 2 & &\\
weapon & 2 & &\\
creature & 2 & &\\
\bottomrule
    \end{tabular}
    \caption{The hyponyms that occur in the train-test split of NMoNLI described in Section~\ref{sec:gen:fine-tune}. The number next to each hyponym corresponds to the number of examples that hyponym occurs in.}
    \label{tab:hyponym}
\end{table}

\subsection{Further Details of Inoculation}\label{app:inoc}

\begin{figure*}[tp]
    \centering
    \begin{tikzpicture}[scale=0.6, line width=2mm]

\begin{axis}[
    title={BERT trained on SNLI},
    every axis plot/.append style={ultra thick},
    xlabel={Number of Examples},
    ylabel={Accuracy},
    xmin=0, xmax=1050,
    ymin=0, ymax=105,
    ytick={0, 10,20,30,40,50,60,70,80,90,100},
    xtick={0,200,400,600,800,1000},
    legend style={at={(0.6, 0.28)}},
    mark size=2pt
]

\addplot[
    color=green,
    mark=square,
    ]
    coordinates {
(100, 90.1)(200, 89.9)(300, 90.5)(500, 83.2)(800,90.5)(1000, 90.2)
    };
\addlegendentry{SNLI Test}

\addplot[
    color=red,
    mark=o,
    ]
    coordinates {
(100, 64.4)(200, 85.1)(300, 37.6)(500, 96.7)(800,90.0)(1000, 94.1)
    };
\addlegendentry{NMoNLI Test}
    
\end{axis}

\end{tikzpicture}
\begin{tikzpicture}[scale=0.6, line width=2mm]

\begin{axis}[
    title={ESIM trained on SNLI},
    every axis plot/.append style={ultra thick},
    xlabel={Number of Examples},
    ylabel={Accuracy},
    xmin=0, xmax=1050,
    ymin=0, ymax=105,
    ytick={0, 10,20,30,40,50,60,70,80,90,100},
    xtick={0,200,400,600,800,1000},
    legend style={at={(0.6, 0.28)}},
    mark size=2pt
]

\addplot[
    color=green,
    mark=square,
    ]
    coordinates {
(100, 86.5)(200, 86.0)(300, 85)(500, 86.3)(800,56.9)(1000, 85.8)
    };
\addlegendentry{SNLI Test}

\addplot[
    color=red,
    mark=o,
    ]
    coordinates {
(100, 39.0)(200, 42.7)(300, 46.0)(500, 42.2)(800,96.2)(1000, 41.8)
    };
\addlegendentry{NMoNLI Test}
    
\end{axis}

\end{tikzpicture}

\begin{tikzpicture}[scale=0.6, line width=2mm]

\begin{axis}[
    title={BiLSTM trained on SNLI},
    every axis plot/.append style={ultra thick},
    xlabel={Number of Examples},
    ylabel={Accuracy},
    xmin=0, xmax=1050,
    ymin=0, ymax=105,
    ytick={0, 10,20,30,40,50,60,70,80,90,100},
    xtick={0,200,400,600,800,1000},
    legend style={at={(0.6, 0.28)}},
    mark size=2pt
]

\addplot[
    color=green,
    mark=square,
    ]
    coordinates {
(100, 71.8)(200, 72.6)(300, 74.6)(500, 77.9)(800,71.0)(1000, 77.2)
    };
\addlegendentry{SNLI Test}

\addplot[
    color=red,
    mark=o,
    ]
    coordinates {
(100, 86)(200, 86.5)(300, 93.5)(500, 90)(800,97)(1000, 90.5)
    };
\addlegendentry{NMoNLI Test}
    
\end{axis}

\end{tikzpicture}
\begin{tikzpicture}[scale=0.6, line width=2mm]

\begin{axis}[
    title={CBOW trained on SNLI},
    every axis plot/.append style={ultra thick},
    xlabel={Number of Examples},
    ylabel={Accuracy},
    xmin=0, xmax=1050,
    ymin=0, ymax=105,
    ytick={0, 10,20,30,40,50,60,70,80,90,100},
    xtick={0,200,400,600,800,1000},
    legend style={at={(0.6, 0.28)}},
    mark size=2pt
]

\addplot[
    color=green,
    mark=square,
    ]
    coordinates {
(100, 65.9)(200, 76.1)(300, 64.6)(500, 69.7)(800,64.6)(1000, 58.4)
    };
\addlegendentry{SNLI Test}

\addplot[
    color=red,
    mark=o,
    ]
    coordinates {
(100, 95.5)(200, 77.5)(300, 90.0)(500, 88.0)(800,88.5)(1000, 90.5)
    };
\addlegendentry{NMoNLI Test}
    
\end{axis}

\end{tikzpicture}



    

    \caption{Inoculation results for our four models performing our systematic generalization task.}
    \label{fig:inoc}
\end{figure*}
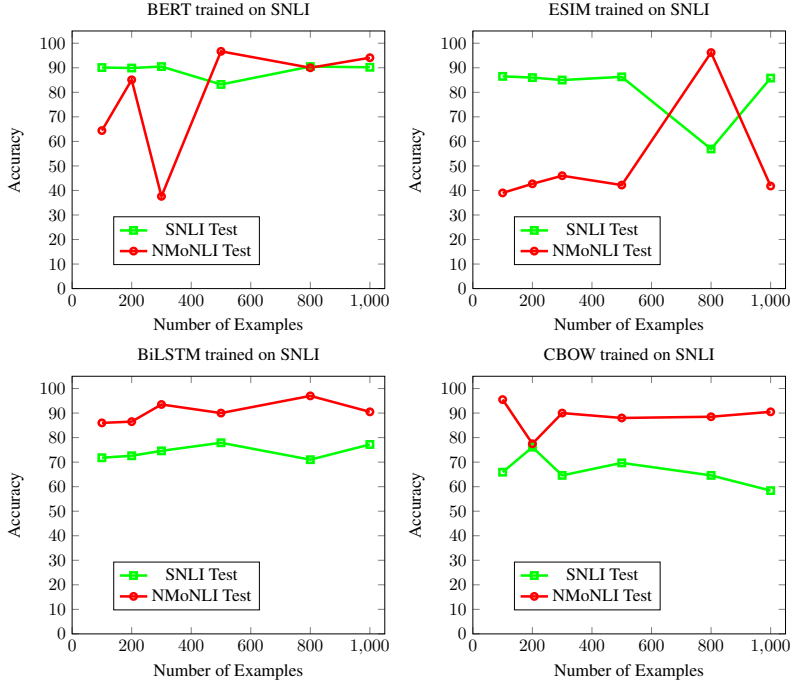

Ideally, a model trained on SNLI that is further trained on NMoNLI will still maintain strong performance on SNLI. We use inoculation by fine-tuning \citep{liu-etal-2019-inoculation} to evaluate models on this ability. In this method, a pretrained model is further fine-tuned on different small amounts of adversarial data while performance on the original dataset and the adversarial dataset is tracked. For each amount of adversarial data, a hyperparameter search is run and the model with the highest average performance on the original dataset and adversarial dataset is selected. Optimizing for the average accuracy is what \citet{richardson2019probing} refer to as \emph{lossless inoculation}, and we perform the same hyperparameter searches that they do. The results of our inoculation experiments are shown in \figref{fig:inoc}. The results in \tabref{tab:adv} under the heading `With \NMoNLI\ fine-tuning' are from the inoculated model with the highest average performance on SNLI test and \NMoNLI\ test.

\subsection{Further Details of Interventions}\label{app:interv}

We say that that BERT mimics the causal dynamics of $\proc{Infer}$ if there is a map $\bertmap$ from \MoNLI\ examples to model-internal vectors in \BERT\ such that the model internal-vectors satisfy the counterfactual claims ascribed to the variable $\lexvar$. Intuitively, $\bertmap$ is a hypothesis about where BERT stores the value of $\lexvar$ for different examples. Our analytic tool for evaluating a map $\bertmap$ is the \tech{interchange intervention}:


Consider inputs $i$ and $j$ and some map from inputs to model-internal vectors $\bertmap$. Suppose that, when BERT is making a prediction for $i$, the vector $\bertmap(i)$ is replaced with the vector $\bertmap(j)$ resulting in output $y$. We say that $y$ is the result of an interchange intervention from $i$ to $j$ under map $\bertmap$ and denote this output as $\interv{i}{j}{\bertmap}$.

In essence, $\interv{i}{j}{\bertmap}$ characterizes the output behavior that results from an experiment where model-internal vectors are interchanged. Recall that $\interfunc{i}{j}{\varfunc}$ describes what output is provided by \proc{Infer} if variables are interchanged. Thus, we can say that \BERT\ \tech{implements} the algorithm \proc{Infer} over a set of examples $S$ if, for all $i,j \in S$, the following equality holds:
\begin{align*}
\interfunc{i}{j}{\varfunc} &= \interv{i}{j}{\bertmap}
\end{align*}
This amounts to observing that the variables in the algorithm and the vectors in the model satisfy the same counterfactual claims.

In the case when $S$ has only two elements $i$ and $j$, we write $\mathcal{X}(i,j)$. For some map $\bertmap$, if $\mathcal{X}(i,j)$ holds for every pair of inputs $i$ and $j$ in \MoNLI, then \BERT\ mimics the causal dynamics of $\proc{Infer}$ on the entirety of \MoNLI.


There are a multitude of possible maps $\bertmap$, and \MoNLI\ has $\approx$2,000 examples, so 7 million interchange interventions must be conducted to verify that BERT mimics the causal dynamics of $\proc{Infer}$ under some map. As such, we must make some assumptions to narrow down our space of possible maps.

When our BERT model processes an example from \MoNLI, it is tokenized as
\[
  e = \langle \CLS, p, \SEP, h, \SEP \rangle
\]
and 12 rows of vector representations are created, so each token is associated with 12 vectors.
In order to efficiently find an appropriate map $\bertmap$, we localize our efforts to the representations created for $\CLS$ and the tokens for the substituted words in the premise and hypothesis, $w_p$ and $w_h$. We additionally assume that every example is mapped to a vector at the same location. This narrows our search to 36 possible maps from inputs in \MoNLI\ to model-internal vectors. For row $r$, we call these $\BERT_{w_p}^{r}$, $\BERT_{w_h}^{r}$, and $\BERT_{\CLS}^{r}$.

Since we must make so many assumptions, we may only be able to find a map that shows $\mathcal{X}(i,j)$ holds for all $i$ and $j$ in some subset of \MoNLI, but not the entirety of \MoNLI\ . Crucially, though, this subset of \MoNLI\ still must contain both lexical relations $\sqsupset$ and $\sqsubset$ for mimicking the causal dynamics of $\proc{Infer}$ to not be vacuous. If one lexical relation is entirely missing from the subset, then none of the interchanges between model vectors will change the output behavior, so there is no guarantee that these vectors play any role in determining output behavior.

As such, we seek the largest subset of \MoNLI\ containing both lexical relations on which BERT implements a modular representation of lexical entailment. To quantify this, we create a graph in which the examples of \MoNLI\ are the nodes and there is an edge between two nodes $n_i$ and $n_j$ if and only if $\mathcal{X}(i,j)$ holds. Cliques in this graph will, in turn, correspond to subsets of \MoNLI\ on which BERT mimics the causal dynamics of \proc{Infer}. We denote the graph for the map $\BERT^r_t$ as $\mathcal{G}_t^r$ for any row $r$ and token $t \in \{\CLS, w_p, w_h\}$.

To see the intuition behind this graph, it is helpful to consider some logically possible scenarios. First, if no examples interchange under our chosen map $\BERT_t^r$, then our graph for that map, $\mathcal{G}_t^r$, will have no edges at all and BERT mimics the causal dynamics of \proc{Infer} on no subset of \MoNLI. Second, if all examples interchange under our chosen map $\BERT_t^r$, then our graph for that map, $\mathcal{G}_t^r$, will be one enormous clique and BERT mimics the causal dynamics of \proc{Infer} on all of \MoNLI.

\begin{figure*}
    \centering
    \input{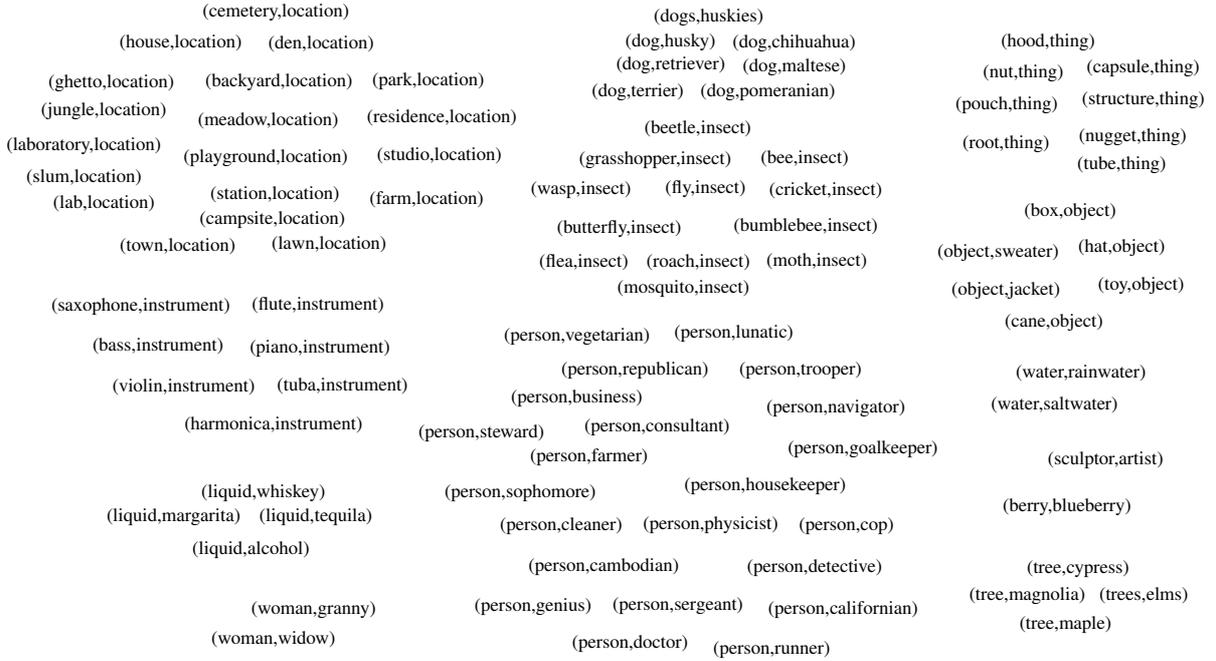}
    \caption{A visualization of the largest subset of \MoNLI\ on which we verified BERT mimics the causal dynamics of \proc{Infer}. This subset contains 98 examples and we display the substituted words in each. The first word in the pair comes from the premise and we cluster word pairs based on hyponyms.}
    \label{fig:cluster}
\end{figure*}


Even with our assumptions restricting us to the 36 maps defined by $\BERT_{w_p}^{r}$, $\BERT_{w_h}^{r}$ and $\BERT^{r}_{\CLS}$, the computational load of performing almost 300 million interchange experiments to construct 36 graphs is too high. Under the constraint of resources, we randomly conducted interchange experiments to partially construct each of the 36 graphs and selected the map whose graph exhibited the most clustering, which was $\BERT_{w_h}^3$.

The problem of finding the largest clique in a graph is NP-complete, so only heuristics are available, but heuristics are fine for the purpose of finding a clique that is large enough. Some edges correspond to interchanges that are causal (the output changes), and some correspond to interchanges that are not causal. To ensure we identify cliques with at least one edge corresponding to a causal interchange, we use the following greedy algorithm: begin with the full graph, and then remove the node with the least number of causal edges until the node with the least number of causal edges has less than $\alpha$, then remove the node with the least number of edges until only a clique remains. We tested $\alpha$ values between 1 and 10 and chose the best results. We seek only cliques that contain a causal edge, because then the subset of \MoNLI\ corresponding to the clique will have both lexical entailment relations represented.


We ran interchange interventions at the location $\BERT^3_{w_h}$ to construct a graph which we partitioned into cliques using our simple, greedy algorithm. We discovered several large disjoint cliques corresponding to subsets of \MoNLI. These cliques had size 98, 63, 47, and 37. We show a visualization of the largest subset on \MoNLI\ containing 98 examples in \figref{fig:cluster}.

To put these results in context, consider a graph with the same number of nodes as the original and edges that were assigned randomly with a 50\% probability. This baseline tells us the level of modularity that would be expected if interchanging a representation randomized the output of the model for its binary classification task. The expected number of cliques  of size $k$ for this graph (2,678 nodes; edge probability of $0.5$) is ${n \choose k} \times 2^{k \choose 2}$. Thus, for $k>20$, the expected number of cliques with $k$ nodes is less than $10^{-8}$.

\end{document}